%% file: main.tex
\documentclass[sigconf,nonacm]{acmart}
\AtBeginDocument{%
  }

\usepackage[utf8]{inputenc} 
\usepackage{booktabs}       
\usepackage{amsfonts}       
\usepackage{nicefrac}       
\usepackage{microtype}      
\usepackage[table]{xcolor}         
\usepackage{xspace}
\usepackage{tablefootnote}
\usepackage{mdframed}
\usepackage{placeins}
\usepackage{multirow}
\usepackage{mathtools}
\usepackage{booktabs}
\usepackage[most]{tcolorbox}
\usepackage{lipsum}
\usepackage{etoolbox}
\usepackage{tabularx}
\usepackage{amsthm}

\usepackage{tikz}
\usepackage{dsfont} 
\usepackage{pifont}
\usepackage{makecell}
\usepackage{subcaption}

\setcopyright{acmlicensed}
\copyrightyear{2018}
\acmYear{2018}
\acmDOI{XXXXXXX.XXXXXXX}
\acmConference[Conference acronym 'XX]{Make sure to enter the correct
  conference title from your rights confirmation email}{June 03--05,
  2018}{Woodstock, NY}
\acmISBN{978-1-4503-XXXX-X/2018/06}

\begin{document}

\input{commands}

\title{QiMeng-CodeV-SVA: Training Specialized LLMs for Hardware Assertion Generation via RTL-Grounded Bidirectional Data Synthesis}

\author{Yutong Wu}
\affiliation{%
  \institution{\SKLP}
  \city{Beijing}
  \country{China}
}
\affiliation{%
  \institution{\UCAS}
  \city{Beijing}
  \country{China}
}
\email{wuyutong22s@ict.ac.cn}

\author{Chenrui Cao}
\affiliation{%
  \institution{\SKLP}
  \city{Beijing}
  \country{China}
}
\affiliation{%
  \institution{\USTC}
  \city{Hefei}
  \state{Anhui}
  \country{China}
}

\author{Pengwei Jin}
\affiliation{%
  \institution{\SKLP}
  \city{Beijing}
  \country{China}
}

\author{Di Huang}
\affiliation{%
  \institution{\SKLP}
  \city{Beijing}
  \country{China}
}

\author{Rui Zhang}
\affiliation{%
  \institution{\SKLP}
  \city{Beijing}
  \country{China}
}

\author{Xishan Zhang}
\affiliation{%
  \institution{\SKLP}
  \city{Beijing}
  \country{China}
}
\affiliation{%
  \institution{\CT}
  \city{Beijing}
  \country{China}
}

\author{Zidong Du}
\affiliation{%
  \institution{\SKLP}
  \city{Beijing}
  \country{China}
}

\author{Qi Guo}
\affiliation{%
  \institution{\SKLP}
  \city{Beijing}
  \country{China}
}

\author{Xing Hu}
\authornote{Corresponding author.}
\affiliation{%
  \institution{\SKLP}
  \city{Beijing}
  \country{China}
}




\input{sections/0_abstract}
\maketitle
\input{sections/1_introduction}

\input{sections/2_related_works}

\input{sections/4_methods}
\input{sections/6_experiments}

\input{sections/7_conclusion}

\input{sections/8_acknowledgement}

\bibliographystyle{ACM-Reference-Format}
\bibliography{references}

\end{document}

%% file: commands.tex
\newcommand{\todo}[1]{{\color{blue}{#1}}}
\newcommand{\di}[1]{{\color{blue}{#1}}}

\newcommand{\modelname}{CodeV-SVA\xspace}
\newcommand{\xname}{Bidirectional Data Synthesis\xspace}

\newcommand{\RR}{DeepSeek-R1\xspace}
\newcommand{\VV}{DeepSeek-V3.1\xspace}
\newcommand{\JG}{JasperGold\xspace}
\newcommand{\HUMAN}{NL2SVA-Human\xspace}
\newcommand{\MACHINE}{NL2SVA-Machine\xspace}
\newcommand{\indicator}[1]{\mathds{1}{#1}}
\definecolor{codegreen}{rgb}{0,0.6,0}
\definecolor{codegray}{rgb}{0.5,0.5,0.5}
\definecolor{codepurple}{rgb}{0.58,0,0.82}
\definecolor{backcolour}{rgb}{0.91,0.91,0.9}
\definecolor{shallowRed}{rgb}{1, 0.8, 0.8}
\definecolor{shallowYellow}{rgb}{1, 0.953, 0.8}
\definecolor{shallowBlue}{rgb}{0.8, 0.8, 1}
\definecolor{orange}{rgb}{1, 0.6, 0}
\definecolor{shallowOrange}{rgb}{1, 0.88, 0.7}

\definecolor{keywordcolor}{rgb}{0.7, 0.1, 0.1}   
\definecolor{commentcolor}{rgb}{0.4, 0.4, 0.4}   
\definecolor{symbolcolor}{rgb}{0.0, 0.1, 0.6}    
\definecolor{sortcolor}{rgb}{0.1, 0.5, 0.1}      
\definecolor{errorcolor}{rgb}{1, 0, 0}           
\definecolor{stringcolor}{rgb}{0.5, 0.3, 0.2}    

\definecolor{leftcolor}{rgb}{0.522, 0.765, 0.863}  
\definecolor{midcolor}{rgb}{0.855, 0.776, 0.812}
\definecolor{rightcolor}{rgb}{0.886, 0.753, 0.596}


\newcommand{\circled}[1]{%
  \tikz[baseline=(char.base)]{
    \node[shape=circle,fill=black,draw,inner sep=1pt, font=\fontsize{8pt}{8pt}\selectfont] (char) {\textcolor{white}{#1}};%
  }%
}

\newcommand{\UCAS}{University of Chinese Academy of Sciences}
\newcommand{\SKLP}{SKL of Processors, Institute of Computing Technology, CAS}
\newcommand{\CT}{Cambricon Technologies}
\newcommand{\SHIC}{Shanghai Innovation Center for Processor Technologies}
\newcommand{\ISCAS}{Intelligent Software Research Center, Institute of Software, CAS}
\newcommand{\ICT}{Institute of Computing Technology, CAS}
\newcommand{\USTC}{University of Science and Technology of China}

%% file: sections/0_abstract.tex
\begin{abstract}

SystemVerilog Assertions (SVAs) are crucial for hardware verification. Recent studies leverage general-purpose LLMs to translate natural language properties to SVAs (NL2SVA), but they perform poorly due to limited data. We propose a data synthesis framework to tackle two challenges: the scarcity of high-quality real-world SVA corpora and the lack of reliable methods to determine NL-SVA semantic equivalence. For the former, large-scale open-source RTLs are used to guide LLMs to generate real-world SVAs; for the latter, bidirectional translation serves as a data selection method. With the synthesized data, we train CodeV-SVA, a series of SVA generation models. Notably, CodeV-SVA-14B achieves 75.8\% on NL2SVA-Human and 84.0\% on NL2SVA-Machine in Func.@1, matching or exceeding advanced LLMs like GPT-5 and DeepSeek-R1.

\end{abstract}

%% file: sections/1_introduction.tex
\section{Introduction}
\label{sec:introduction}

Assertion-based hardware formal verification plays a vital role in the digital hardware design flow~\cite{ABV_Servey}. Based on natural-language specifications and Register Transfer Level (RTL) code, verification engineers need to formulate temporal logical assertions, namely
SystemVerilog Assertions (SVAs)~\cite{mehta2020systemverilog}, and employ simulation or formal verification (FV) tools (e.g., Cadence JasperGold) to ensure that the RTL implementation satisfies the specified constraints corresponding to the SVAs.

Given the substantial manual effort and expertise required to craft high-quality SVAs, the automatic generation of SVAs has emerged as an important research area. Classical rule-based methods \cite{HARM, ARTmine} mine SVAs from simulation traces, but they depend on the availability of golden RTL designs and are thus hard to apply to real-world verification scenarios.
More recent work focuses on employing large language models (LLMs) to analyze specifications and RTL, first formulating verification properties in natural language and then translating those properties into corresponding SVAs \cite{AssertLLM, AssertionForge, DeepAssert, AssertCoder, AssertGen}. The common weakness of these methods is their usage of general-purpose LLMs (e.g., \VV \cite{deepseekv3}) for the natural-language-to-SVA (NL2SVA) translation. Due to the lack of background knowledge, general-purpose LLMs like \VV tend to underperform on this specialized translation task \cite{FVEval, AssertionsbyLLMs} (see Table~\ref{tab:main_results}). In addition, advanced LLMs such as GPT-5~\cite{GPT-5} and DeepSeek-R1~\cite{deepseekr1} either fail to meet proprietary requirements of hardware companies or impose very high deployment costs. Therefore, developing LLMs that are both accurate for SVA generation and affordable to deploy is essential.

However, training SVA LLMs is hampered by the lack of high-quality training data~\cite{VERT}, which stems from two main challenges.
\textbf{(1) The scarcity of high-quality SVA corpora in the wild.} Publicly available human-written SVAs mostly appear in textbooks and a handful of open-source repositories, but these sources are limited in scale. For example, Hybrid-NL2SVA~\cite{Hybrid-NL2SVA} contains only 4,070 SVAs from textbooks; from the DeepCircuitX~\cite{deepcircuitx} dataset, we can only extract 5,638 SVAs in over 4K open-source repositories; 
By contrast, open datasets of RTL code (e.g., DeepCircuitX~\cite{deepcircuitx} and CodeV~\cite{CodeV}) cover on the order of $10^5$ NL-Verilog pairs, a scale and that far exceed those of SVA data. 
\textbf{(2) The lack of reliable methods to determine semantic equivalence between NL properties and model-generated SVAs.} LLM-based data synthesis pipelines typically rely on post-selection to filter low-quality samples \cite{data_selection_survey}, but NL-SVA pairs are hard to validate automatically: Simply using formal verification tools to check the SVA under RTL is insufficient because trivial or vacuous assertions (e.g., \texttt{assert property (1'b1)}) will pass against any RTL, yet show no alignment with the NL description \cite{FVEval}; Another approach, LLM-as-a-judge~\cite{surveyllmasajudge}, which prompts an LLM to judge whether the NL and the generated SVA are consistent, struggles with NL ambiguity and the subtle syntax of SVA (e.g., operator precedence, see Figure~\ref{fig:bidirectional_case_study}).


\begin{figure}[t]
  \centering
\includegraphics[width=0.8\linewidth]{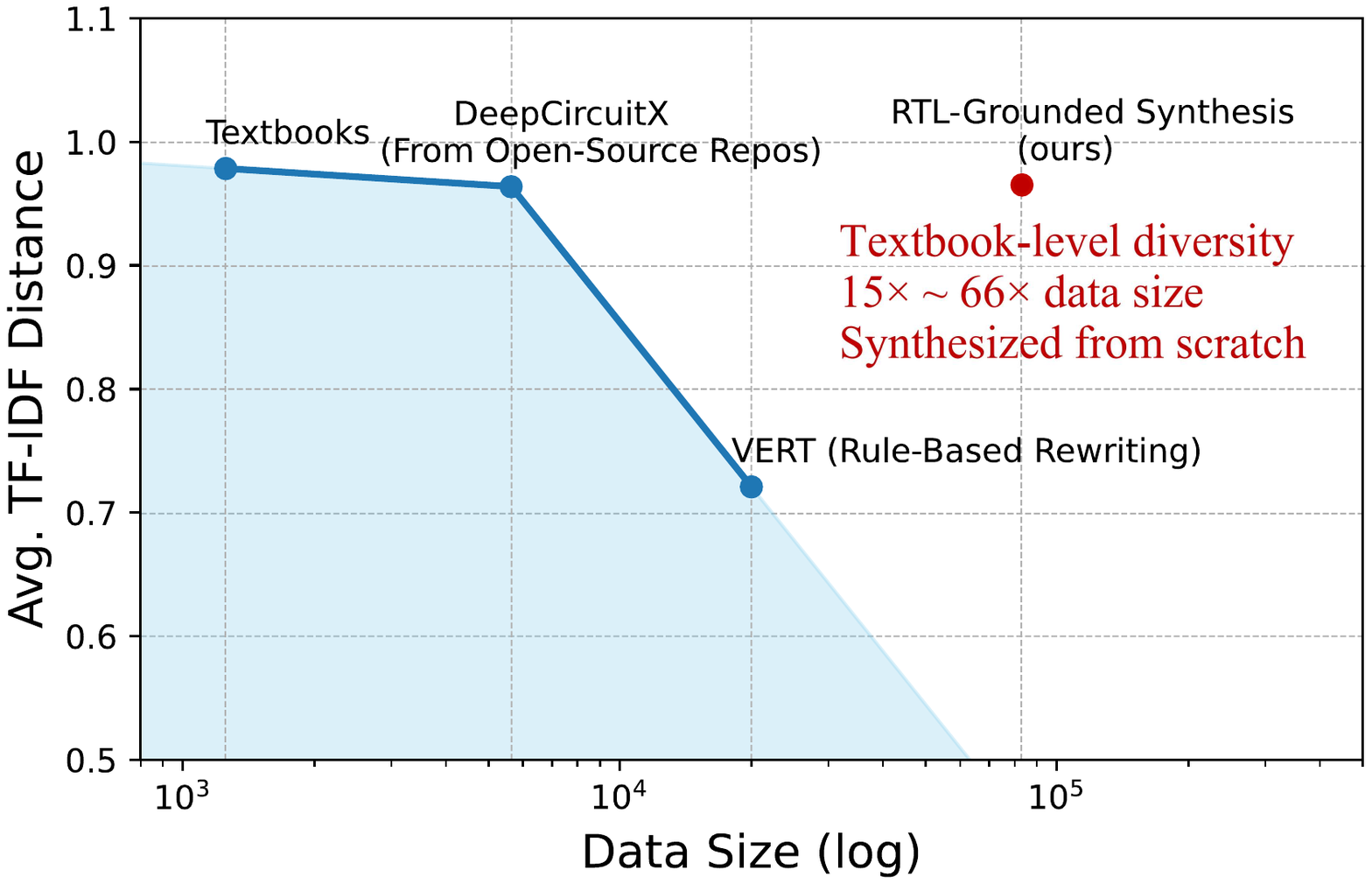}
\caption{The relationship between data size and average 3-gram TF-IDF distance of different SVA sources.}
\label{fig:tfidf}
\end{figure}

To tackle these challenges, we begin with two key observations: \textbf{(1) LLMs can build upon open-source RTL code as design-under-tests (DUTs) to synthesize a large amount of high-quality SVAs, thereby alleviating data scarcity.} To validate this, we analyze the SVAs from textbooks and open-source repositories, SVAs automatically generated by rule-based rewriting in prior work VERT~\cite{VERT}, and RTL-grounded. We measure both the scale and diversity (using the 3-gram TF-IDF distance~\cite{tfidf} as the metric) of them. In Figure~\ref{fig:tfidf}, as data size grows, the diversity of the rule-based method drops significantly, while the RTL-grounded approach maintains diversity close to that of open-source repositories, demonstrating its stronger data augmentation capability. \textbf{(2) Both NL2SVA and SVA2NL translation incur information loss. Consequently, if an SVA, after being converted to NL by LLMs and then back to SVA (bidirectional translation), remains logically equivalent to the original one, it is highly probable that no information loss has occurred, indicating that the SVA and NL are well-aligned.}
We randomly select 50 NL-SVA pairs from our synthetic dataset, and employ human experts to evaluate the accuracy both before and after the selection via bidirectional translation. The results indicate that the accuracy improves from 68\% to 96\%, showing bidirectional translation can serve as an effective selection method. We also show how bidirectional translation identifies subtle errors in Figure~\ref{fig:bidirectional_case_study}. The original SVA can cheat both formal verification and LLM-as-a-judge, but it cannot pass the equivalence check after bidirectional translation.

\input{figures/1_introduction/bidirectional_case_study}

\begin{figure*}[t]
  \centering
\includegraphics[width=0.8\linewidth]{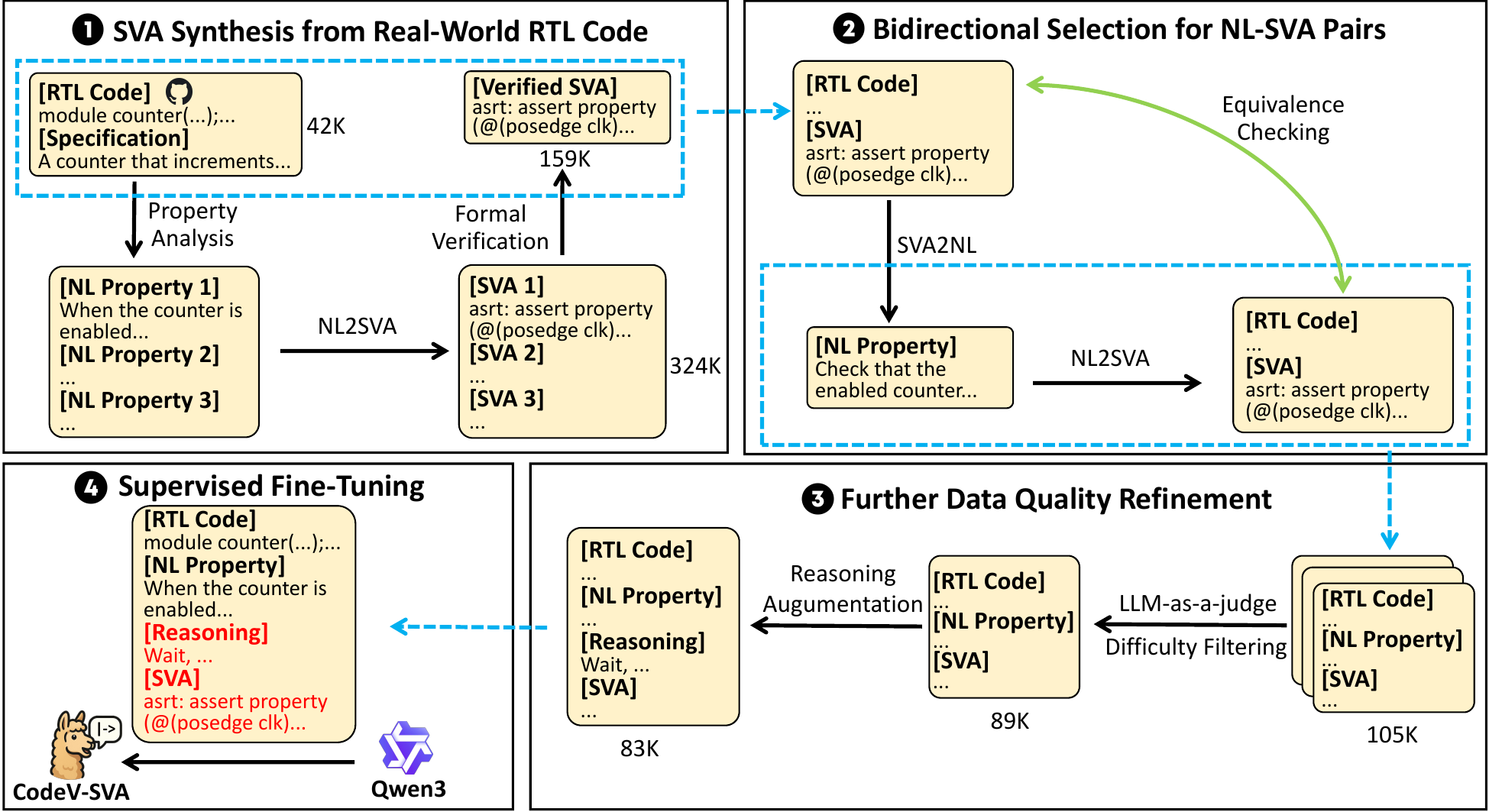}
\caption{\textbf{The overview of our data synthesis and training framework.}}
\label{fig:method}
\end{figure*}

Based on these observations, we propose an efficient data synthesis framework for NL2SVA tasks. Specifically, our process begins with a collection of open-source RTL designs. For each design, we first employ a general-purpose LLM to generate multiple NL verification properties and their SVA counterparts. They are subsequently filtered by a formal verification tool, resulting in a large-scale, high-quality seed SVA dataset. Next, we perform bidirectional translation to further align the NL-SVA pairs. Each SVA is first translated into NL and then back into SVA by the LLM. Formal verification tools are then used to check the logical equivalence between the original and re-translated SVAs. Only the equivalent SVAs and their corresponding NLs are retained. Finally, with further difficulty and LLM-as-a-judge selection, followed by reasoning enhancement, we obtain a high-quality NL2SVA dataset. The overview of our data synthesis framework is shown in Figure~\ref{fig:method}.

To show the effectiveness of our data synthesis method, we develop \modelname, a series of fine-tuned LLMs for NL2SVA. Specifically, with the synthesized data, we perform supervised fine-tuning (SFT) on open-source models, including Qwen3 8B / 14B \cite{qwen3}. We evaluate them on the mainstream NL2SVA benchmark, FVEval-NL2SVA \cite{FVEval}. The results show that \modelname-14B achieves 75.8\% on \HUMAN and 84.0\% on \MACHINE in Func.@1, matching or surpassing SOTA general-purpose LLMs such as GPT-5 and \RR, while requiring minimal deployment resources. We plan to open-source our dataset, models, and training pipeline.

%% file: figures/1_introduction/bidirectional_case_study.tex
\begin{figure}[h]
    \centering
    \begin{tcolorbox}[
      boxrule=0.5pt,     
      arc=2pt,            
      left=4pt, right=4pt,
      top=3pt, bottom=3pt,
      boxsep=1pt,       
      colback=white,
      colframe=black,
      fonttitle=\small,
      title=A Case Study of Bidirectional Data Selection
    ]
    
\textbf{Original SVA:}

\begin{lstlisting}[language=verilog,basicstyle=\small\ttfamily]
asrt_term_complement: assert property (
    @(posedge i_clk) disable iff (tb_reset)
    ctrl_comp |-> (term == (~mux_out + 1))
    and !ctrl_comp |-> (term == mux_out)
);
\end{lstlisting}

\textbf{Formal Verification: \textcolor{codegreen}{Passed}.}

\hfill

\textbf{NL translated from SVA:}

The \texttt{term} value must be the two's complement of \texttt{mux\_out} when \texttt{ctrl\_comp} is high, and equal to \texttt{mux\_out} when \texttt{ctrl\_comp} is low.

\textbf{LLM-as-a-judge: \textcolor{codegreen}{Passed}.}

\hfill

\textbf{New SVA re-translated from NL:}

\begin{lstlisting}[language=verilog,basicstyle=\small\ttfamily,escapechar=?]
asrt_term_complement: assert property (
    @(posedge i_clk) disable iff (tb_reset)
    ?{\colorbox{shallowRed}{(}}?ctrl_comp |-> (term == (~mux_out + 1))?{\colorbox{shallowRed}{)}}?
    and ?{\colorbox{shallowRed}{(}}?!ctrl_comp |-> (term == mux_out)?{\colorbox{shallowRed}{)}}?
);
\end{lstlisting}

\textbf{Equivalent Checking: \textcolor{red}{Failed}.}

\textbf{Remark}: In SVA syntax, the \textbf{and} operator has higher precedence than the implication (\texttt{|->}), which makes the original SVA a meaningless tautology that can pass formal verification under any RTL code. Due to the limited knowledge, this subtle error cannot be identified by LLMs. However, with bidirectional translation, we regenerate a new SVA that is not logically equivalent to the original one, thereby detecting this erroneous data.
\end{tcolorbox}
\vspace{-3mm}
\caption{An example of bidirectional data selection.}
\label{fig:bidirectional_case_study}
\end{figure}

%% file: sections/2_related_works.tex
\section{Related Work}
\label{sec:related_work}

Using LLMs to assist hardware verification is a promising application of LLMs for EDA~\cite{LLM4EDA}, encompassing LLM-based testbench generation~\cite{AutoBench,CorrectBench}, LLM-assisted RTL bug localization and repair~\cite{FVDebug,AssertFix}, LLM-aided Universal Verification Methodology (UVM)~\cite{UVLLM,UVLLM2}, and LLM for SVA generation~\cite{FVEval, AssertLLM, AssertionForge, VERT, Hybrid-NL2SVA}, which is the primary focus of this paper. 

Most LLM-based SVA generation approaches employ existing general-purpose LLMs to generate natural-language properties and SVAs: AssertLLM~\cite{AssertLLM} uses the Retrieval Augmented Generation (RAG) to improve SVA generation flow; AssertionForge~\cite{AssertionForge} constructs Knowledge Graphs (KGs) to help LLMs better understand the relationships between specifications and RTL; DeepAssert~\cite{DeepAssert} guides LLMs to generate SVAs by analyzing the invocation relationships between modules. In contrast, we use synthesized data to train specialized LLMs rather than employ a general model, thereby further improving the accuracy and efficiency of SVA generation.

On the other hand, VERT \cite{VERT} collects SVAs and RTL code from open-source repositories to finetune LLMs for generating syntactically correct and verifiable SVAs. However, its training data lacks NL properties, making the LLM unable to generate SVAs from NL intents directly provided by human engineers. Hybrid-NL2SVA \cite{Hybrid-NL2SVA} collects 4K SVAs from textbooks and uses LLMs to generate NL explanations as paired training data, but the small size of the training data limits its effectiveness. We expand the dataset to 83K with RTL-grounded synthesis and improve data quality by bidirectional selection. Moreover, since Hybrid-NL2SVA's datasets are not publicly available, we are unable to make a performance comparison.

%% file: sections/4_methods.tex
\section{Methods}
\label{sec:methods}

\noindent\textbf{Preliminaries.} We follow the problem definition and evaluation protocol for the NL2SVA translation task in FVEval~\cite{FVEval}. Specifically, we take RTL code $c$ and a natural-language verification property $x$ as the input of an LLM $\mathcal{M}$, and then use a formal verification tool (e.g., \JG) to check the logical equivalence (denoted as $\sim$) between the model-generated SVA $\mathcal{M}(c,x)$ and the ground-truth SVA $\hat{y}$ given by human experts. Formally, the NL2SVA translation of $x$ by $\mathcal{M}$ is \textbf{functionally correct} if and only if $\mathcal{M}(c,x) \sim \hat{y}$.

\noindent\textbf{Overview.} This section presents the data synthesis and training framework of \modelname in detail, which consists of 4 stages:
\circled{1} \textbf{SVA Synthesis from Real-World RTL Code.} To alleviate the scarcity of SVA data, we use a general-purpose LLM to analyze a large scale of open-source RTL code and generate corresponding NL properties and SVAs, followed by filtering out verifiable SVAs by a formal verification tool.
\circled{2} \textbf{Bidirectional Selection for NL-SVA Pairs.} Since a verifiable SVA may not fully capture the NL semantics (e.g., \texttt{assert property (1'b1)}), we align NL-SVA via bidirectional translation: SVAs are back-translated to NLs and then re-translated into new SVAs. Only the instances whose newly generated SVAs are proven equivalent to the original one by formal tools are retained. 
\circled{3} \textbf{Further Data Quality Refinement.} We integrate human-expert insights into an LLM-as-a-judge selection and remove trivial data pairs by a weaker general-purpose LLM to enhance training effectiveness. Then, we use a strong reasoning LLM, \RR, to augment the dataset with reasoning trajectories. 
\circled{4} \textbf{Supervised Fine-Tuning.} With the synthetic NL2SVA dataset, we fine-tune general-purpose LLMs to obtain \modelname.
\vspace{-4mm}
\subsection{SVA Synthesis from Real-World RTL Code}
\label{sec:method_sva_synthesis}

\textbf{(1) Open Source RTL Code Curation.} Unlike SVA data, open-source RTL code is sufficiently abundant. We choose the CodeV dataset \cite{CodeV} for SVA synthesis, which contains 165K RTL code (denoted as $\{c_i\}$) collected from GitHub and corresponding specification (denoted as $\{s_i\}$) generated by LLMs. To collect the code more suitable for verification scenarios, we use Yosys~\cite{yosys} to analyze the signals in each RTL and filter out 42K instances $\{(c_i, s_i)\}$ with clock and reset signals, as they typically involve temporal constraints and can thus inspire the model to generate more complex SVAs.

\input{figures/4_methods/prompt_and_few_shots}

\textbf{(2) Natural-Language Verification Property Analysis.} To improve the diversity and accuracy of the synthesis process, we first employ \VV to decompose the specification $\{s_i\}$ into multiple independent properties $\{x_{ij}\}$ (see Figure~\ref{fig:prompts}a for prompts and example), helping the model better focus on each subtask. There are 324K generated properties in total.

\textbf{(3) SVA Generation and Verification.} 
We again feed each synthetic NL property $x_{ij}$ with its corresponding RTL code $c_i$ into \VV, which generates an SVA $y_{ij}$ that can be used to check whether $x_{ij}$ holds. To perform an initial screening of high-quality SVAs, we use \JG to identify the formally verified SVAs $\{y^*_{ij}\}$ under the RTL code $c_i$, yielding our original SVA corpora $\mathcal C = \{(\{y^*_{ij}\}_{j=1}^{m_i}, c_{i})\}$, where $m_i$ is the number of the verified SVAs for $c_i$. Finally, we obtain 159K SVA instances for the subsequent synthesis of NL-SVA paired data.
\vspace{-1mm}
\subsection{Bidirectional Selection for NL-SVA Pairs}
\label{sec:method_bidirectional}
Although SVA $y^*_{ij}$ passes formal verification, it does not imply that it fully reflects the semantics of NL $x_{ij}$, since the RTL code can also satisfy weaker SVAs, and training on such data would significantly degrade the model's performance (see Table~\ref{tab:ablation_data_selection}). We first attempt to include SVA tutorials \cite{SVA_DOC} in the prompt to assist the model, but the performance shows no considerable improvement (see Table \ref{tab:doc_vs_selection}), so we choose to conduct data selection. 

\input{tables/4_methods/doc_vs_selection}

Our main selection method is bidirectional translation (see Sec. \ref{sec:introduction}), which proceeds in two steps:

\textbf{(1) SVA-to-NL-to-SVA Translation.} The SVA $y^*_{ij}$ is first translated to NL property $x^*_{ij}$ by \VV, and then translated back to another SVA $y'_{ij}$. For SVA-to-NL, we use few-shot examples (Figure~\ref{fig:prompts}b) to guide the model to generate high-level NL properties rather than describe signal relationships directly. For NL-to-SVA, we incorporate the signals extracted from the original SVA into the prompt as hints to reduce the uncertainty of LLM outputs.

\textbf{(2) Data Filtering Based on Formal Equivalence Checking.} Since the misalignment may arise in any of the two translation directions, if the regenerated SVA is logically equivalent to the original, the corresponding NL-SVA pair is likely to be consistent. Following FVEval~\cite{FVEval}, we use \JG to formally verify the logical equivalence between two SVAs. More formally, we construct the aligned dataset $\mathcal{D}$ as follow:
\begin{align*}
\mathcal{D} = \{ (c_i, x^*_{ij}, y'_{ij}) \mid x^*_{ij} = \mathcal{M}^{-1}(c_i, y^*_{ij}), y'_{ij} = \mathcal{M}(c_i, x^*_{ij}), y'_{ij} \sim y^*_{ij}\},
\end{align*}
where $\mathcal{M}$ and $\mathcal{M}^{-1}$ is the LLM's NL-to-SVA and SVA-to-NL process, and $(\cdot \sim \cdot)$ is the equivalence verification. After filtering, we obtain 105K NL-SVA pairs with corresponding RTL code.

\subsection{Further Data Quality Refinement}
\label{sec:method_data_refinement}
Besides bidirectional data selection, we apply additional general techniques to further refine the data quality, including:

\textbf{(1) LLM-as-a-judge integrated with expert priors.} We employ human experts to analyze LLM errors that bidirectional translation could not identify, and categorize the errors into 4 types: logical misalignment, signal inconsistency, RTL misunderstanding, and mapping the NL property to the wrong SVA object. We ask \VV to select the NL-SVA pairs without these errors.

\textbf{(2) Difficulty Filtering via a Weaker LLM.} To improve training efficiency, we use Qwen3-8B, a general-purpose LLM with weak SVA generation ability (see Table~\ref{tab:main_results}), to generate 5 SVAs $\{y_{ijk}\}_{k=1}^5$ for each NL $x^*_{ij}$ and remove the instance where all SVAs are equivalent to $y'_{ij}$, thereby filtering trivial data points.

\textbf{(3) Reasoning Trajectory Augmentation.} Following OpenAI o1~\cite{openai-o1}, to fully leverage the model’s reasoning capability during NL2SVA, we use a reasoning-enhanced LLM to augment the dataset with long reasoning. Specifically, we feed each NL $x^*_{ij}$ to \RR and obtain the answer SVA $y''_{ij}$ with reasoning trajectory $r_{ij}$. We retain only the data whose $y''_{ij} \sim y'_{ij}$, since the correct final answer often implies a correct reasoning trajectory~\cite{star}. 

\subsection{Supervised Fine-Tuning}

Finally, we obtain an NL2SVA training dataset of 83K instances (i.e., $\{(c_i, x_{ij}^*, r_{ij}, y''_{ij})\}$). To evaluate the quality of the dataset, we fine-tune general-purpose LLMs on it. We treat the RTL code $c_i$ as DUT, and concatenate it with the NL $x_{ij}^*$ as the input of the model. For training label $\tilde{y}_{ij}$, we follow the \RR format to place the reasoning trajectory between special tokens, followed by the SVA answer, i.e., $\tilde{y}_{ij}= \texttt{<think>}r_{ij}\texttt{</think>}y''_{ij}$. Formally, we minimize the following objective loss:
\begin{equation*}
L_{\text{SFT}}\left(\theta\right) = -\sum_{i=1}^{N} \sum_{j=1}^{M_i} \sum_{t=1}^{T_{ij}} \log P(\tilde{y}^{(t)}_{ij} \mid \tilde{y}^{(<t)}_{ij},c_{i}||x^*_{ij}; \theta),
\end{equation*}
where $N$ denotes the number of RTL codes, $M_i$ denotes the number of training data associated with the $i$-th RTL, $T_{ij}$ denotes the number of tokens in $\tilde{y}_{ij}$, $\tilde{y}^{(t)}_{ij}$ and $\tilde{y}^{(<t)}_{ij}$ denote the $t$-th token of $\tilde{y}_{ij}$ and its preceding tokens, and $\theta$ denotes the model parameters. We optimize the log probability of the output tokens with respect to the training labels, thereby improving the model's NL2SVA performance.

%% file: figures/4_methods/prompt_and_few_shots.tex
\begin{figure}[h]
    \centering

    \input{figures/4_methods/property_analysis_prompt}
    \input{figures/4_methods/sva2nl_prompt}
    \vspace{-4mm}
    \caption{Concise prompts and examples of NL verification property analysis (Sec. \ref{sec:method_sva_synthesis}) and SVA2NL (Sec. \ref{sec:method_bidirectional}).}
    \label{fig:prompts}
\end{figure}

%% file: figures/4_methods/property_analysis_prompt.tex
\begin{tcolorbox}[
  boxrule=0.5pt,     
  arc=2pt,            
  left=4pt, right=4pt,
  top=3pt, bottom=3pt,
  boxsep=1pt,       
  colback=white,
  colframe=black,
  fonttitle=\small,
  title=(a) The Prompt and an Example of Property Analysis
]
    
You are given a hardware specification together with its Verilog implementation. You need to provide a list of verification properties for this design.

Example:

\#\#\# Specification

...the program counter in Verilog that increments its address by 1 every clock cycle when the enable signal is high.

\#\#\# Code Implementation

\begin{lstlisting}[language=verilog,basicstyle=\small\ttfamily]
...
always @(posedge clock or negedge rst) begin
    ...
    else begin
        if(en) pc_addr <= pc_addr+1;
        else pc_addr <= pc_addr;
    end
...
\end{lstlisting}

[Response]

Property 1: When disabled, the counter must hold its value.
    
\end{tcolorbox}

%% file: figures/4_methods/sva2nl_prompt.tex
    \begin{tcolorbox}[
      boxrule=0.5pt,     
      arc=2pt,            
      left=4pt, right=4pt,
      top=3pt, bottom=3pt,
      boxsep=1pt,       
      colback=white,
      colframe=black,
      fonttitle=\small,
      title=(b) The Prompt and an Example of SVA2NL
    ]
    
You are given a hardware design in Verilog and a SVA that verifies some property of the design. Your task is to describe the properties being checked by the SVA in natural language.

Example:

\#\#\# Hardware Design

\begin{lstlisting}[language=verilog,basicstyle=\small\ttfamily]
...
always @(posedge clk or posedge reset) begin
    ...
    if (cmd_valid == 1) begin
        cmd_reg = cmd;
        busy <= 1;
    end
...
\end{lstlisting}

\#\#\# SVA

\begin{lstlisting}[language=verilog,basicstyle=\small\ttfamily]
asrt: assert property (
    @(posedge clk) disable iff (tb_reset)
    (cmd_valid && !busy) |=> busy
);
\end{lstlisting}

[Response]

When a command is issued while the controller is idle, the controller must become busy in the next cycle.

    \end{tcolorbox}

%% file: tables/4_methods/doc_vs_selection.tex
\begin{table}[h]
\centering
\setlength{\tabcolsep}{1mm}
\fontsize{9}{11}\selectfont
\caption{The performance (\%) of \RR with and without tutorial prompts. See Sec. \ref{sec:exp_settings} for details of benchmarks.}
\begin{tabular}{l|cc|cc}
\toprule
 \multirow{2}{*}{Method}     & \multicolumn{2}{c|}{\HUMAN} & \multicolumn{2}{c}{\MACHINE} \\
 & Func.@1     & Func.@16    & Func.@1      & Func.@16     \\
\midrule
w/o Tutorial  & \textbf{74.6}   & \textbf{90.3}   & 81.0                 & 93.3                \\
w/ Tutorial   & 71.5            & 90.0            & \textbf{83.7}        & \textbf{93.7}       \\
\bottomrule
\end{tabular}
\label{tab:doc_vs_selection}
\end{table}

%% file: sections/6_experiments.tex
\section{Experiments}
\label{sec:Experiments}

In this section, we present our implementation details (Sec. \ref{sec:implementation_details}). We conduct a series of experiments (settings in Sec. \ref{sec:exp_settings}) to compare the performance between \modelname and other LLMs (Sec. \ref{sec:main_results}), the contributions of each component in our method (Sec. \ref{sec:ablation_study}) and the role of \modelname in an end-to-end verification workflow (Sec. \ref{sec:end2end}).

\subsection{Implementation Details}
\label{sec:implementation_details}

In data synthesis, we set the temperature to 0.8 for NL property generation and SVA2NL translation to increase diversity. We use greedy decoding for NL2SVA to ensure the accuracy of the generated SVAs. In SFT, we use the LlamaFactory~\cite{llamafactory} framework to fine-tune Qwen3 8B / 14B. The models are trained for 2 epochs with a learning rate of 2e-5 and a batch size of 128. The context length is 32,768. The training of 8B and 14B models takes 8 and 12 hours on 8 H800-80G GPUs. We use \JG 2023.12 for evaluation.

\subsection{Experimental Settings}
\label{sec:exp_settings}

\noindent\textbf{Benchmarks.} To show the performance of \modelname, we evaluate it on \HUMAN and \MACHINE of FVEval \cite{FVEval}. \HUMAN covers expert-written ground-truth SVAs and NL descriptions. \MACHINE randomly generates a batch of SVAs by manual rules, then uses LLMs to generate the NL descriptions. We employ human experts to recheck the benchmarks, correcting or removing erroneous tests. We perform 13-gram decontamination \cite{GPT-3} on the training datasets to avoid contamination of benchmarks. 

\noindent\textbf{Metrics.} We use full functional correctness in FVEval~\cite{FVEval} as the metric, i.e., the model-generated SVA holds if and only if the ground-truth SVA holds. We use the pass@$k$~\cite{codex} metric to measure whether the model can generate at least one correct SVA within $k$ attempts, denoted as Func.@$k$:
\begin{align*}
\text{Func.@$k$} &:= \mathop{\mathbb{E}}_{\text{Problems}} \left[ 1 - \frac{{\binom{n-c}{k}}} {\binom{n}{k}} \right],
\end{align*}
where $n$ is the number of samples for a problem and $c$ is the number of correct SVAs generated by LLMs. In this work, we set $n = 32$.

\noindent\textbf{Baselines.} We compare \modelname with advanced general-purpose LLMs, including \RR (2025-05-28) \cite{deepseekr1}, GPT-5 (2025-08-07) \cite{GPT-5}, DeepSeek-V3.1 (2025-08-21) \cite{deepseekv3} and GPT-4o (2024-11-20) \cite{GPT-4o}. We also evaluate the specialized RTL generation LLMs, including RTLCoder-Deepseek-v1.1~\cite{RTLCoder} and CodeV-R1-Qwen-7B~\cite{CodeV-R1}, as their RTL generation capabilities may transfer to SVA generation. We also evaluate Qwen3-8B and 14B \cite{qwen3} to show the performance improvement from training data. All models are evaluated with a temperature of 0.8 and a \texttt{top\_p} of 0.95.

\subsection{Main Results}
\label{sec:main_results}

\input{tables/5_experiments/main_results}

The evaluation results of \modelname on FVEval-NL2SVA are shown in Table \ref{tab:main_results}. The results demonstrate that \modelname's performance leads most general-purpose LLMs:

\textbf{(1) Our \modelname-14B model establishes new
state-of-the-art results on the Func.@1 scores of both benchmarks, surpassing general-purpose LLMs as well as specialized RTL generation LLMs, demonstrating the effectiveness of our data synthesis and training pipeline.} Notably, \modelname-14B surpasses or matches its teacher model \RR-671B and expensive proprietary LLM GPT-5 across all Func.@$k$ of the two NL2SVA benchmarks, providing a clear computational and cost efficiency advantage. Furthermore, both the 8B and 14B models push beyond previous models' upper bound on Func.@32 of \MACHINE.

\textbf{(2) Both \modelname-8B and 14B achieve substantial performance gains over their base models, Qwen3-8B and 14B.} In particular, \modelname-8B achieves 39.7\% and 37.4\% improvements over Qwen3-8B in Func.@1 of \HUMAN and \MACHINE, respectively, demonstrating that \textbf{high-quality synthesized data} is the key to improving SVA generation capabilities of open-source general-purpose LLMs. 

\subsection{Ablation Studies}
\label{sec:ablation_study}

To investigate the contribution of each component in the data synthesis framework of \modelname, we conduct a series of ablation studies, including ablations on the sources of SVA in the training dataset (Table \ref{tab:ablation_data_source}) and on the data refinement and selection methods (Table \ref{tab:ablation_data_selection}). We use Qwen3-8B as the base model in all ablations.

\textbf{(1) LLM-synthesized SVAs are high-quality training labels, which can contribute to better performance than collecting or rewriting existing SVAs from open-source repositories.} We compare the SVAs generated by LLMs (Section \ref{sec:method_sva_synthesis}) with the SVAs collected or rule-based rewritten from open-source repositories in prior work, including DeepCircuitX \cite{deepcircuitx} and VERT \cite{VERT}. Specifically, we use \VV to perform SVA2NL (Section \ref{sec:method_bidirectional}) on the open-source SVAs $\{y_i\}$ with RTL code $\{c_i\}$ to obtain the NL properties $\{x_i\}$, then train Qwen3-8B on $\{(c_i, x_i, y_i)\}$. We compare the resulting LLMs with LLMs trained on the same amount of synthetic data of our method (random 5K). The results (Table~\ref{tab:ablation_data_source}) show that training LLMs on open-source SVAs leads to a significant degradation in performance, which is likely due to the variable data quality across these repositories, which makes training unstable.

\input{tables/5_experiments/ablation_study_data_source}

\input{tables/5_experiments/ablation_study_data_selection}

\textbf{(2) In data refinement (Section \ref{sec:method_data_refinement}), reasoning augmentation provides a clear performance boost for the model. Difficulty filtering and LLM-as-a-judge serve complementary roles that reduce data size and improve training efficiency}. In Table~\ref{tab:ablation_data_selection}, removing the reasoning trajectories (line 2), the performance drops across all metrics on both benchmarks, highlighting the role of the reasoning paradigm in the NL2SVA task. Using difficulty filtering (line 3) and LLM-as-a-judge (line 4), we remove 15\% low-quality samples and achieve a modest improvement in the model's Func.@1 score.

\textbf{(3) Bidirectional data selection (Section~\ref{sec:method_bidirectional}) yields the most substantial performance gain for NL2SVA-Human (12.3\% in Func.@1, line 4 vs. line 5) among all components of our data synthesis framework, fully demonstrating the method's effectiveness in enhancing data quality.}     Furthermore, by reducing the amount of data by 34\%, the bidirectional selection method greatly improves training efficiency.

\textbf{(4) With only SVAs that pass formal verification under RTL code (Section~\ref{sec:method_sva_synthesis}), the model achieves superior performance with less than half of the training data (line 5 vs. line 6).} Although there is a slight performance drop on \MACHINE, which may be attributed to the distribution gap between the rule-based synthesized benchmark and real-world SVAs, it is easy to recover via subsequent data refinement.
The whole ablation studies show that the selection stages in our method continuously improve the performance while reducing the data size.

\subsection{\modelname in End-to-End Verification}

\label{sec:end2end}

\input{tables/5_experiments/assertionforge}

We evaluate the effectiveness of \modelname within a fully automated end-to-end verification framework: given the specification document of a hardware design as input, LLMs autonomously analyze the design constraints and generate the corresponding SVAs. Specifically, we modify the AssertionForge framework~\cite{AssertionForge} by dividing it into two components: (1) Spec2NL: General-purpose LLMs (e.g., GPT-4o) are used to analyze the specifications and RTL code, generating NL properties; (2) NL2SVA: The NL properties are translated into corresponding SVAs by \modelname.  We keep the same framework in AssertionForge, modify its prompts for our model, and evaluate it on 5 RTL designs~\cite{opencores, AssertLLM} in the original paper. We keep GPT-4o and \RR in the Spec2NL task, while we choose \modelname for the NL2SVA task. Then we count the total number of generated SVAs (\#SVA), the number of syntax correct SVAs (\#Sync), and the number of SVAs that pass formal verification (\#Proven) as metrics (COI coverage in AssertionForge's paper is already saturated; thus, \modelname only brings improvement to OPENMSP430, and the remaining designs achieve nearly 100\% coverage).

\textbf{Given the same NL plans, \modelname generates significantly more syntax correct and verifiable SVAs during the NL2SVA stage (Table \ref{tab:assertionforge}).} Notably, on OPENMSP430, a complex design with a 129-page specification document and 29 RTL files, \modelname-8B generates far more verifiable SVAs than GPT-4o (2.5$\times$) and \RR (3.5$\times$), demonstrating its advantage over general-purpose LLMs in an end-to-end verification framework.

%% file: tables/5_experiments/main_results.tex
\begin{table}[h]
\centering
\setlength{\tabcolsep}{0.9mm}
\caption{The evaluation results (\%) of \modelname and other LLMs. ``F@$k$'' refers to Func.@$k$ score (Section \ref{sec:exp_settings}).}
\begin{tabular}{l|ccc|ccc}
\toprule
\multirow{2}{*}{Model}     & \multicolumn{3}{c|}{\HUMAN} & \multicolumn{3}{c}{\MACHINE} \\
                    & F@1               & F@16            & F@32                    & F@1               & F@16              & F@32   \\
\midrule
\multicolumn{7}{l}{(\textit{Advanced General-Purpose Models})} \\
\RR-671B                 & \underline{74.6}      & \textbf{90.3}       & \underline{90.4}            & 81.0                  & 93.3                  & 94.3 \\
GPT-5               & 71.8                  & \underline{90.2}    & \textbf{92.7}               & 81.8                  & 93.2                  & 94.3 \\
\VV-671B                 & 63.1                  & 81.4                & 84.9                        & \underline{83.8}      & 92.9                  & 93.6 \\
GPT-4o              & 64.1                  & 75.2                & 78.1                        & 68.5                  & 81.3                  & 83.7 \\
\midrule
\multicolumn{7}{l}{(\textit{Specialized RTL Generation Models})} \\
RTLCoder-DS-v1.1-6.7B  & 25.9 & 58.8 & 65.8 & 21.7 & 54.8 & 60.8 \\
CodeV-R1-Qwen-7B      & 25.2 & 55.8 & 61.6 & 37.4 & 76.6 & 83.0 \\
\midrule
\multicolumn{7}{l}{(\textit{Open-Source Foundation Models})} \\
Qwen3-8B            & 32.3                  & 71.6                & 74.0                        & 46.1                  & 88.0                  & 90.5 \\
Qwen3-14B           & 61.6                  & 86.1                & 87.7                        & 75.3                  & 92.7                  & 94.3 \\
\midrule
\rowcolor[rgb]{0.925,0.925,0.925} \multicolumn{7}{l}{(\textit{Ours})} \\
\rowcolor[rgb]{0.925,0.925,0.925} \modelname-8B       & 72.0       & 88.8       & \underline{90.4}      & 83.5     & \textbf{96.3}      & \textbf{97.2} \\
\rowcolor[rgb]{0.925,0.925,0.925} \modelname-14B      & \textbf{75.8}       & 89.4       & \underline{90.4}      & \textbf{84.0}     & \underline{94.9}      & \underline{95.8} \\

\bottomrule
\end{tabular}
\label{tab:main_results}
\end{table}

%% file: tables/5_experiments/ablation_study_data_source.tex
\begin{table}[h]
\centering
\setlength{\tabcolsep}{0.8mm}
\caption{The performance (\%) comparison between different SVA sources in training data. Synthesized: LLM-synthesized SVAs. DeepCircuitX: SVAs in open-source repositories. VERT: Rewriting open-source SVAs with manual rules.}
\begin{tabular}{l|c|ccc|ccc}
\toprule
\multirow{2}{*}{SVA Source}  & Data   & \multicolumn{3}{c|}{\HUMAN} & \multicolumn{3}{c}{\MACHINE} \\
                            &  Size                               & F@1               & F@16            & F@32            & F@1              & F@16              & F@32   \\
\midrule
\rowcolor[rgb]{0.925,0.925,0.925} Synthesized (ours)  & 5K  & \textbf{55.4} & \textbf{84.0} & \textbf{86.3} & \textbf{75.7} & \textbf{94.0} & \textbf{95.1} \\
DeepCircuitX                      & 5K  & \underline{22.3}     & \underline{59.0}            & \underline{64.4}            & \underline{39.2}             & \underline{58.8}              & \underline{61.1} \\
VERT & 20K & 1.9 & 8.8 & 11.0 & 6.4 & 16.0 & 19.1 \\
\bottomrule
\end{tabular}
\label{tab:ablation_data_source}
\end{table}

%% file: tables/5_experiments/ablation_study_data_selection.tex
\begin{table}[h]
\centering
\setlength{\tabcolsep}{0.9mm}
\caption{The performance (\%) of removing each component in data refinement (Sec. \ref{sec:method_data_refinement}) and selection (Sec. \ref{sec:method_bidirectional} and \ref{sec:method_sva_synthesis}). R: Reasoning trajectories; D:  Difficulty filtering; J: LLM-as-a-judge; B: Bidirectional data selection; V: Formal verification under RTL code.}
\begin{tabular}{l|c|ccc|ccc}
\toprule
\multirow{2}{*}{Method}  & Data  & \multicolumn{3}{c|}{\HUMAN} & \multicolumn{3}{c}{\MACHINE} \\
                                                      & Size   & F@1               & F@16            & F@32            & F@1              & F@16              & F@32   \\
\midrule
\rowcolor[rgb]{0.925,0.925,0.925} \modelname (ours) &  83K   & \textbf{72.0}     & \textbf{88.8}   & \textbf{90.4}   & \textbf{83.5}    & \textbf{96.3}     & \textbf{97.2} \\
\midrule
\multicolumn{8}{l}{(\textit{Ablation on data refinement})} \\
w/o R                                                  &  89K   & \underline{63.9}              & 80.0            & \underline{82.2}            & \underline{81.1}             & 91.2              & 92.2 \\
w/o R, D                                               &  97K   & 62.2              & \underline{80.4}            & \underline{82.2}            & \underline{81.1}             & 91.6              & 92.6 \\
w/o R, D, J                                            &  105K  & 63.5              & 80.1            & \underline{82.2}            & 77.2             & 91.5              & 92.6 \\
\midrule
\multicolumn{8}{l}{(\textit{Ablation on data selection})} \\
w/o R, D, J, B                                         &  159K  & 51.2              & 76.4            &	80.8            & 76.5             & 92.1              & 92.9 \\
w/o R, D, J, B, V                                      &  324K  & 44.1              & 73.5            & 77.4            & 78.8             & \underline{93.0}              & \underline{94.0} \\
\bottomrule
\end{tabular}
\label{tab:ablation_data_selection}
\end{table}

%% file: tables/5_experiments/assertionforge.tex
\definecolor{lightgray}{rgb}{0.925,0.925,0.925}
\begin{table}[h]
\centering
\caption{Evaluation results of \modelname on AssertionForge.}
\vspace{-1mm}
\label{tab:results}
\begin{tabular}{llccc}
\toprule
Spec2NL & NL2SVA & \#SVA & \#SynC &\#Proven \\
\hline 
\multicolumn{5}{c}{\textsc{APB}} \\
\hline
\multirow{2}{*}{GPT-4o}  & GPT-4o                    & 436 & 377 & 106 \\
 & \cellcolor{lightgray}\modelname-8B (ours)      & \cellcolor{lightgray}\textbf{442} & \cellcolor{lightgray}\textbf{384} & \cellcolor{lightgray}\textbf{180} \\
\hline
\multirow{2}{*}{\RR}     & \RR                       & 515 & 281 & 122 \\
 & \cellcolor{lightgray}\modelname-8B (ours)      & \cellcolor{lightgray}\textbf{529} & \cellcolor{lightgray}\textbf{466} & \cellcolor{lightgray}\textbf{211}  \\
\hline

\multicolumn{5}{c}{\textsc{ETHMAC}} \\
\hline
\multirow{2}{*}{GPT-4o}  & GPT-4o         & 720 & 565 & 49  \\
 & \cellcolor{lightgray}\modelname-8B (ours)  & \cellcolor{lightgray}\textbf{721} & \cellcolor{lightgray}\textbf{579} & \cellcolor{lightgray}\textbf{76}  \\
\hline
\multirow{2}{*}{\RR}     & \RR            & \textbf{665} & 371 & 75   \\
 & \cellcolor{lightgray}\modelname-8B (ours)  & \cellcolor{lightgray}660 & \cellcolor{lightgray}\textbf{532} & \cellcolor{lightgray}\textbf{84}  \\
\hline

\multicolumn{5}{c}{\textsc{OPENMSP430}} \\
\hline
\multirow{2}{*}{GPT-4o}  & GPT-4o         & 1013 & 777 & 196 \\
 & \cellcolor{lightgray}\modelname-8B (ours) & \cellcolor{lightgray}\textbf{1044} & \cellcolor{lightgray}\textbf{859} & \cellcolor{lightgray}\textbf{484} \\
\hline
\multirow{2}{*}{\RR}     & \RR            & \textbf{1222} & 496 & 144 \\
 & \cellcolor{lightgray}\modelname-8B (ours) & \cellcolor{lightgray}1215 & \cellcolor{lightgray}\textbf{917} & \cellcolor{lightgray}\textbf{502}  \\
\hline

\multicolumn{5}{c}{\textsc{SOCKIT}} \\
\hline
\multirow{2}{*}{GPT-4o}  &  GPT-4o        & \textbf{251} & 173 & 31 \\
 & \cellcolor{lightgray}\modelname-8B (ours)  & \cellcolor{lightgray}\textbf{251} & \cellcolor{lightgray}\textbf{202} & \cellcolor{lightgray}\textbf{49}  \\
\hline
\multirow{2}{*}{\RR}     & \RR            & 232 & 122 & 40 \\
 & \cellcolor{lightgray}\modelname-8B (ours)  & \cellcolor{lightgray}\textbf{247} & \cellcolor{lightgray}\textbf{178} & \cellcolor{lightgray}\textbf{58} \\
\hline

\multicolumn{5}{c}{\textsc{UART}} \\
\hline
\multirow{2}{*}{GPT-4o}  & GPT-4o         & 265 & 186 & 54  \\
 & \cellcolor{lightgray}\modelname-8B (ours)  & \cellcolor{lightgray}\textbf{270} & \cellcolor{lightgray}\textbf{196} & \cellcolor{lightgray}\textbf{73}  \\
\hline
\multirow{2}{*}{\RR}     & \RR            & 258 & 91  & 40 \\
 & \cellcolor{lightgray}\modelname-8B (ours)  & \cellcolor{lightgray}\textbf{266} & \cellcolor{lightgray}\textbf{185} & \cellcolor{lightgray}\textbf{70}  \\
\bottomrule

\end{tabular}
\label{tab:assertionforge}
\end{table}

%% file: sections/7_conclusion.tex
\section{Conclusion}
\label{sec:Conclusion}

In this paper, we propose \modelname, a data synthesis and model training pipeline for SVA generation. This pipeline alleviates the scarcity of high-quality SVA data by synthesizing SVAs from real-world RTL code, and further improves data quality by filtering model-generated NL-SVA data pairs using a bidirectional translation method, followed by data refinement. We train \modelname-8B and 14B with this pipeline and evaluate them on \HUMAN and \MACHINE benchmarks of FVEval. Experimental results show that \modelname significantly outperforms general-purpose LLMs on NL2SVA tasks, demonstrating the effectiveness of our data synthesis and training pipeline.

%% file: sections/8_acknowledgement.tex
\section{Acknowledgements}
\label{sec:Acknowledgements}

This work is partially supported by the NSF of China (Grants No.62402477, 62341411), Strategic Priority Research Program of the Chinese Academy of Sciences (Grants No.XDB0660300, XDB0660301, XDB0660302, XDB0660200, XDB0660201, XDB0660202), and Youth Innovation Promotion Association CAS.